\documentclass{article}
\usepackage[utf8]{inputenc}
\usepackage{amsmath}
\usepackage{graphicx}
\usepackage[colorinlistoftodos]{todonotes}
\usepackage[colorlinks=true, allcolors=blue]{hyperref}
\usepackage{indentfirst}
\usepackage[a4paper,top=3cm,bottom=2cm,left=3cm,right=3cm,marginparwidth=1.75cm]{geometry} 
\usepackage{titlesec}
\usepackage{hyperref}
\usepackage{booktabs} 
\usepackage{float}

\titleclass{\subsubsubsection}{straight}[\subsection]

\newcounter{subsubsubsection}[subsubsection]
\renewcommand\thesubsubsubsection{\thesubsubsection.\arabic{subsubsubsection}}

\titleformat{\subsubsubsection}
  {\normalfont\normalsize\bfseries}{\thesubsubsubsection}{1em}{}
\titlespacing*{\subsubsubsection}
{0pt}{3.25ex plus 1ex minus .2ex}{1.5ex plus .2ex}

\title{Applying Dimensionality Reduction as Precursor to LSTM-CNN Models for Classifying Imagery and Motor Signals in ECoG-Based BCIs}
\author{Soham Bafana}
\begin{document}
\maketitle

\section*{Abstract}

Motor impairments, frequently caused by neurological incidents like strokes or traumatic brain injuries, present substantial obstacles in rehabilitation therapy. This research aims to elevate the field by optimizing motor imagery classification algorithms within Brain-Computer Interfaces (BCIs). By improving the efficiency of BCIs, we offer a novel approach that holds significant promise for enhancing motor rehabilitation outcomes. Utilizing unsupervised techniques for dimensionality reduction, namely Uniform Manifold Approximation and Projection (UMAP) coupled with K-Nearest Neighbors (KNN), we evaluate the necessity of employing supervised methods such as Long Short-Term Memory (LSTM) and Convolutional Neural Networks (CNNs) for classification tasks. Importantly, participants who exhibited high KNN scores following UMAP dimensionality reduction also achieved high accuracy in supervised deep learning (DL) models. Due to individualized model requirements and massive neural training data, dimensionality reduction becomes an effective preprocessing step that minimizes the need for extensive data labeling and supervised deep learning techniques. This approach has significant implications not only for targeted therapies in motor dysfunction but also for addressing regulatory, safety, and reliability concerns in the rapidly evolving BCI field.\newline
\newline
\textbf{Keywords}: BCI, Dimesionality Reduction, CNN, LSTM, ECoG, UMAP

\section{Introduction}

Motor Skill Impairment (MSI), prevalent in conditions like strokes, traumatic brain injuries, and neuromuscular pathologies, results in significant deficits in both gross and fine motor functionalities \cite{miaoBCIBasedRehabilitationStroke2020}. A major focus in neural rehabilitation research is motor imagery, a cognitive process entailing simulated muscular movement without overt motor execution \cite{macintyreMotorImageryPerformance2018}. This mechanism, deeply rooted in motor planning and sensorimotor proprioception, mirrors actual motor activity, suggesting its potential utility in MSI-targeted rehabilitative strategies \cite{miaoBCIBasedRehabilitationStroke2020}. However, its complex nature poses challenges for integration, notably in brain-computer interfaces (BCIs).

BCIs play a crucial role in understanding motor execution function and developing potential therapies for motor control-related disorders. Thus, datasets of brain recordings are key for developing novel BCI systems \cite{thompsonComputationalLimitsDeep2020}. In particular, motor imagery and execution datasets have been extensively studied due to the availability of public datasets and the growing interest in data-driven approaches \cite{dae-cheolgwonReviewPublicMotor2023}. These datasets have been used to investigate various aspects of motor control, such as motor execution, motor imagery, and imagery-based online feedback. Ultimately, datasets are key for the implementation of deep learning (DL) strategies. 

ECoG and EEG (electroencephalogram) datasets have seen a gradual increase in recent years, with studies focusing on various aspects of brain activity, such as attention, directional orientation, and spatial awareness. In novel literature, deep learning techniques have been applied to classify, predict, and analyze ECoG and EEG datasets across neurological studies \cite{liDeepLearningEEG2020}. In particular, common methodologies include preprocessing to reduce artifact interference and signal processing to isolate the range of frequency when extracting particular features and brain waves. One can then convert the extracted features to particular commands and use them in a BCI based system \cite{torresEEGBasedBCIEmotion2020}.

Arguably, ECoG (Electrocorticography) data possesses many advantages over other neuroimaging systems. ECoG involves the placement of a grid of electrodes directly on the exposed surface of the brain, typically during neurosurgical procedures for epilepsy or tumor removal \cite{Hill2012-pe}. This direct contact allows for the collection of neural data with higher spatial and temporal resolution compared to non-invasive methods like EEG (Electroencephalography). In addition, ECoG systems are not susceptible to contamination from muscle movements and eye blinks, issues that can regularly impair the quality of other neuroimaging methodologies. Leuthardt et al. \cite{leuthardtBrainComputerInterface2004} found that ECoG-based BCIs could provide individuals with motor skill impairments (MSI) non-muscular communication and control options that are more powerful than EEG-based BCIs and more stable than electrodes implanted deeper into the brain. ECoG BCIs offer not only a larger data pool but also a safer avenue for invasive recording \cite{volkovaDecodingMovementElectrocorticographic2019}. ECoG-based studies have paved the way for innovative applications, such as improving epilepsy surgery through deep learning-assisted functional mapping, thereby mitigating the risks associated with Electrical Stimulation Mapping (ESM) \cite{raviprakashDeepLearningProvides2020}. Moreover, classifiers like Support Vector Machine and Gaussian Naive Bayes have been employed to decode semantic information from ECoG signals, offering intriguing insights into the brain's cognitive processes \cite{wangDecodingSemanticInformation2011}.

Simultaneously, the Machine Learning (ML) landscape has undergone transformative changes, particularly in computer vision and natural language processing \cite{krizhevskyImageNetClassificationDeep2017, devlinBERTPretrainingDeep2018}. These strides have been predominantly fueled by the growth in dataset sizes and computational capabilities \cite{thompsonComputationalLimitsDeep2020}. One pivotal technique that has emerged in handling high-dimensional data is dimensionality reduction. Among the methods available, Uniform Manifold Approximation and Projection (UMAP) stands out for its versatility in visualizing and understanding complex datasets \cite{mcinnesUMAPUniformManifold2018}. Deep learning techniques, like Convolutional Neural Networks (CNNs) and Recurrent Neural Networks (RNNs) featuring Long Short-Term Memory (LSTM) units, have been integrated into BCIs to process high-dimensional data effectively. CNNs excel in tasks requiring grid-like data structures, such as image recognition \cite{kimConvolutionalNeuralNetwork2017}. In contrast, LSTMs are tailored for sequential data, making them apt for decoding ECoG signals \cite{duDecodingECoGSignal2018, garza-ulloaDeepLearningModels2022}. 

However, the BCI domain faces unique challenges, primarily due to the restricted availability of neural signal databases. These limitations stem from a range of factors including experimental constraints, clinical considerations, and hardware limitations \cite{sliwowskiDeepLearningECoG2022}. To address these issues, we propose a supervised deep learning approach augmented by dimensionality reduction and K-Nearest Neighbors (KNN) algorithms. In particular, we aim to understand the degree to which UMAP accuracy correlates with the accuracy in DL paradigms. Furthermore, we illustrate how LSTM and CNN architectures can be employed for complex signal prediction tasks.

In conclusion, deep learning technologies are essential in bridging the gap between computational systems and neural interfaces. As we delve deeper into this integration, it becomes increasingly vital to explore how dimensionality reduction and deep learning paradigms can interact to enhance the analysis of ECoG datasets.

\section{Materials and Methods}

In our study, we employed a multi-step approach to investigate the efficacy of dimensionality reduction techniques and supervised learning models in the classification of electrocorticography (ECoG) signals for Brain-Computer Interfaces (BCI). Our methodology included data preprocessing, dimensionality reduction using UMAP, and model building with Convolutional Neural Networks (CNN) and Long Short-Term Memory networks (LSTM). The first step in this process was to find a suitable dataset. 

\subsection{Dataset Description}

The Miller ECoG Motor Execution and Imagery Dataset \cite{millerCorrectionCorticalActivity2010} was comprised of electrocortiography (ECoG) recordings from patients.  This dataset contains seven sets of data, each corresponding to a unique patient, with data segregated into real and imagery categories. Each of the fourteen unique recordings contains a time-series voltage signal `v', sampled at 1000 Hz from different brain regions (stored as `s-rate'). These recordings also include event markers (`t-on', `t-off'), which indicate the timing of stimuli, and stimulus identifiers (`stim-id'), which distinguish whether the stimulus is tongue- or hand-based, each represented by 11 or 12 respectively. The data also contains the respective locations of recordings in the hemisphere, locs, gyrus and Broadman Area. 

In the initial Miller paper, the magnitude of imagery-induced cortical change exceeded that of actual movement during a BCI-ECoG task of moving a cursor on a screen \cite{millerCorrectionCorticalActivity2010}. Thus, further classification and prediction of these imagery induced ECoG signals would provide more information on building a BCI system for related signals. 

\subsection{Data Cleaning}

Upon an initial examination of the dataset, several challenges were identified. The voltage data, for instance, is organized as a two-dimensional array representing instances and brain regions. However, the patients differed in the number of brain regions examined, resulting in datasets of unequal dimensions. To rectify this issue, two separate strategies were implemented: padding the array with null values when required, or truncating each set of voltages to the smallest uniform length. Both strategies proved instrumental in pre-processing the data for further analysis \cite{harris2020array}.

Furthermore, the voltage signals ('V') were originally stored as raw signals. Before modeling, we passed the raw signals through a high-pass filter at 50 Hz, followed by a low-pass filter at 10 Hz, and then normalized via absolute value and squaring. This transformation process enabled the conversion of the original signal data into a time-series format based on the signal's approximate power, which was important for visualization and further analysis. The data was then cleaned to ensure that only the 2000 voltage readings after every stimulus remained \cite{scikit-learn}.

\subsection{Exploratory Data Analysis}

Initially, we investigated in the frequency domain

\textbf{Fourier Analysis:} We computed the Fourier Transform of the ECoG signals to isolate the dominant frequencies. This analysis disclosed different spectral characteristics for real and imaginary movements. Interestingly, the peaks for both real and imagery data occurred at around the same frequencies (200, 300, 400), suggesting a level of correlation between the two.


\textbf{Power Spectral Density Analysis:} We determined the Power Spectral Density (PSD) for each Fourier-transformed signal to identify the distribution of power over various frequencies.

\begin{figure}[H]
\centering
\includegraphics[width=16cm]{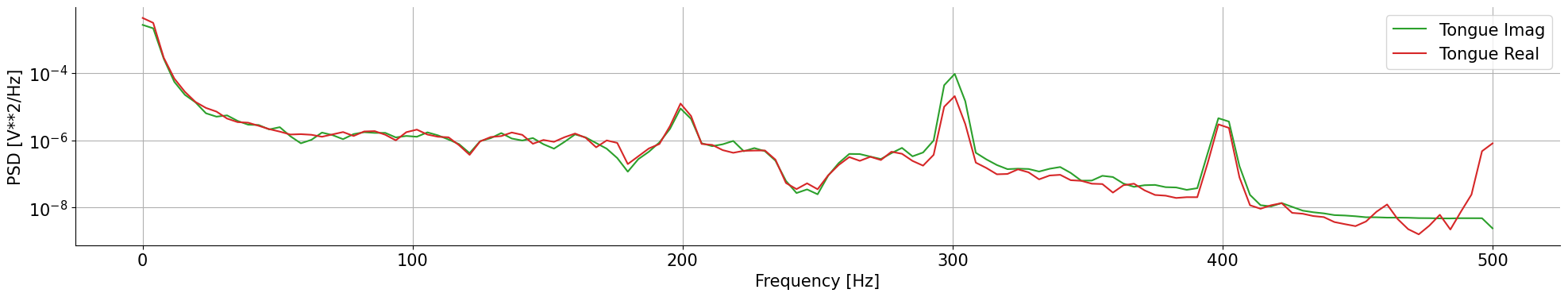}
\caption{Power Spectral Density of the Tongue Stimulus}
\label{fig:PsdTongue}
\end{figure}

\begin{figure}[H]
\centering
\includegraphics[width=16cm]{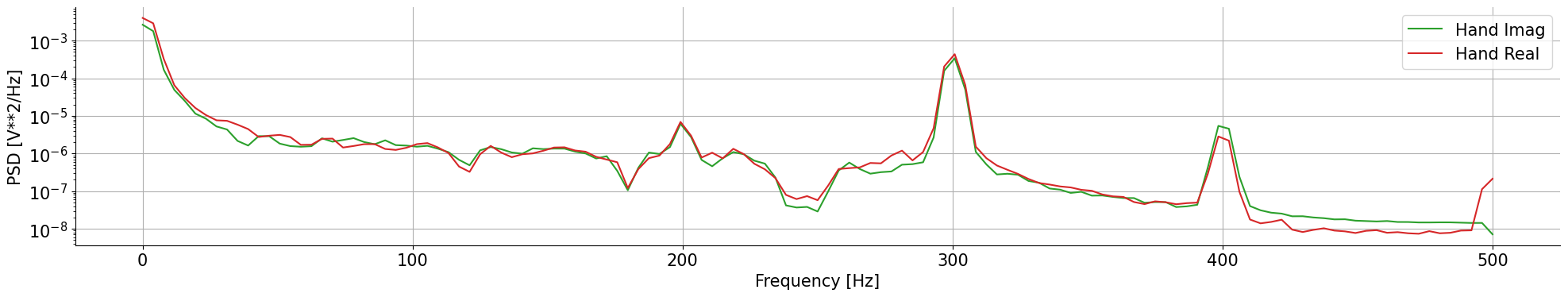}
\caption{Power Spectral Density of the Hand Stimulus}
\label{fig:PsdHand}
\end{figure}

The PSD values calculated for both sets of data exhibited strong similarity, indicating a lack of a clear distinction between real and imagery data based on power distribution across frequencies. To further quantify the correlation between the signals, we calculated the coherence between pairs of ECoG signals as a function of frequency.

\begin{figure}[H]
\centering
\includegraphics[width=16cm]{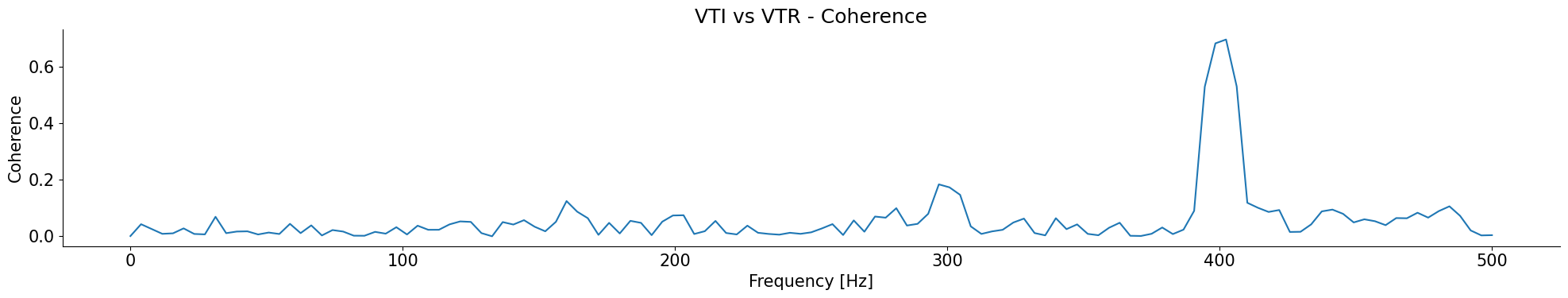}
\caption{Coherence of the Tongue Imagery-Induced and Real Frequency}
\label{fig:CoherenceTongue}
\end{figure}

\begin{figure}[H]
\centering
\includegraphics[width=16cm]{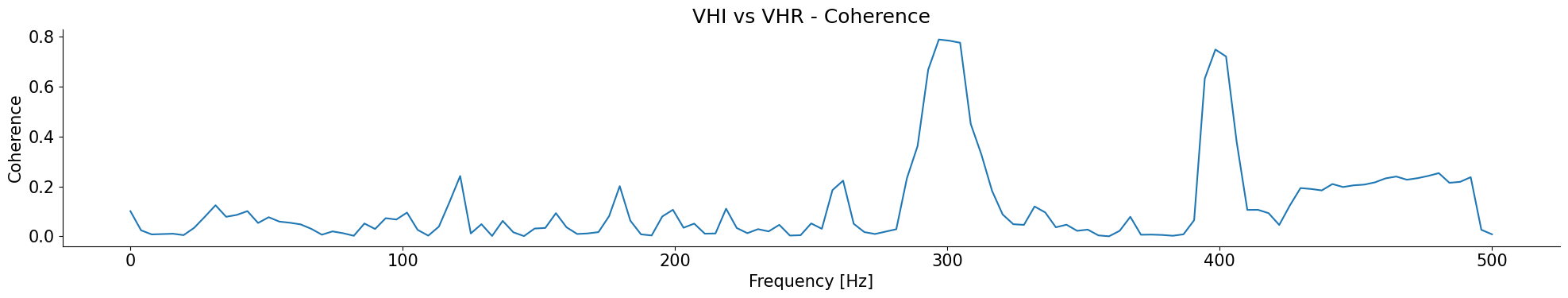}
\caption{Coherence of the Hand Imagery-Induced and Real Frequency}
\label{fig:CoherenceHand}
\end{figure}

The coherence between the real and imagery-induced datasets was very similar, with a few spikes that indicated only minor differences. This further underscores the similarity between imagery-induced motor execution and actual motor execution in the frequency domain. This processed data provided an indication of the differences between real and imagery-induced data.

In conclusion, these findings affirm that imagery and real data are closely interrelated, exhibiting significant similarities in both frequency and time-series analysis. This confluence of similarities and differences offers intriguing insights into brain activity during motor tasks and highlights the potential of ECoG signals in the development of brain-computer interfaces. 

\subsection{Modeling}

\subsubsection{Dimensionality Reduction and KNN}

Dimensionality reduction techniques, such as Uniform Manifold Approximation and Projection (UMAP), are essential for transforming high-dimensional data into a more manageable, low-dimensional space while preserving its significant characteristics. Widely applied in fields like signal processing, speech recognition, neuroinformatics, and bioinformatics, UMAP offers a robust framework grounded in manifold learning and topological data analysis. It has been especially useful for visualizing and understanding complex, high-dimensional datasets \cite{mcinnesUMAPUniformManifold2018}. In our specific context, we leverage UMAP to perform dimensionality reduction on stimuli data, aiming to uncover inherent features that facilitate the automatic sorting and prediction of ECoG data in an unsupervised manner. The UMAP algorithm achieves this by initially constructing a k-neighbour graph from the high-dimensional dataset. In this graph, nodes symbolize data points and are linked to their k nearest neighbors, with weighted edges that capture the distances between nodes. Subsequently, a "fuzzy simplicial set" transformation is applied to preserve both the local and global structures inherent in the data.

Addressing the known asymmetry in k-neighbour graphs, UMAP implements a unification process to enhance symmetry. The method then stipulates an objective function targeting the conservation of the data's topological structure during the reduction process, emphasizing the retention of relative distances and densities within the newly defined space. Ultimately, through a gradient descent approach, UMAP identifies a lower-dimensional projection that optimizes the objective function, resulting in a representation that maintains the structural and relational characteristics of the original high-dimensional data \cite{mcinnesUMAPUniformManifold2018}.

The phenomenon of diverse resolutions in signals across different individuals is highlighted by \cite{huSubjectSeparationNetwork2023}, which further complicates cross-individual classification. In response, our methodological approach concentrates on individualized classification tailored for Brain-Computer Interface (BCI) systems.

\subsubsection{Neural Networks}

Deep learning methodologies, including Convolutional Neural Networks (CNNs) and Recurrent Neural Networks (RNNs) with Long Short-Term Memory (LSTM) units, have proven instrumental in analyzing high-dimensional data in Brain-Computer Interfaces (BCIs). CNNs, specialized for grid-like data structures such as images, comprise multiple layers like convolutional, activation (ReLU), and pooling layers. These layers facilitate the hierarchical detection of patterns, ranging from simple lines to complex structures \cite{kimConvolutionalNeuralNetwork2017}. On the other hand, LSTMs excel in handling sequential data, owing to their capability to learn long-term dependencies. They employ 'gates' to regulate the flow of information, thereby maintaining a stable training gradient \cite{garza-ulloaDeepLearningModels2022}. Notably, LSTMs have been applied to decode and classify ECoG signals \cite{duDecodingECoGSignal2018}. Moreover, the transformation of BCI signals into image form offers innovative avenues for predicting motor imagery impulses \cite{vicentea.lomelin-ibarraMotorImageryAnalysis2022}.

In our research, we extend the application of these deep learning architectures, focusing on CNNs and hybrid CNN-LSTM networks, to further the state-of-the-art in dimensionality reduction techniques for ECoG datasets. Specifically, we engaged with data from participants that had previously shown poor performance in dimensionality reduction paradigms. For our test models, we selected data from participant 2, who exhibited the lowest accuracy in earlier modeling efforts.

Our approach offers a multi-dimensional analysis that captures both spatial and temporal aspects of the brain's electrical activity. The inclusion of CNNs empowers our architecture to efficiently capture spatial features, especially beneficial for identifying localized patterns in ECoG data from multiple electrodes. CNNs achieve this by focusing on local spatial dependencies, thereby enhancing the interpretability of these spatial features for subsequent analytical tasks \cite{liDeepLearningEEG2020}. Complementing the spatial capture capability of CNNs, LSTMs focus on the temporal dynamics of the ECoG signals. They provide a mechanism for understanding how brain states evolve over time, thus offering a more comprehensive representation of the neural activities. This is particularly crucial for BCI applications where understanding the sequential dependencies of brain signals can provide more accurate and reliable outputs \cite{wangUnsupervisedDecodingLongTerm2016}. Additionally, the hybrid CNN-LSTM architecture allows for feature fusion by merging the extracted spatial and temporal features in the fully connected layers. This not only broadens the feature set but also significantly improves the model's classification accuracy. The architecture also employs middle-layer feature extraction through a flatten layer, enabling the capture of complex features that are often overlooked but can be vital for enhancing classification performance \cite{liMotorImageryEEG2022}. Finally, we attempted a process called Fine-tuning, to further refine our model. Fine-tuning involves adjusting the pre-trained weights in our network by continuing the training process on our specific ECoG data set. This allows the model to better adapt to the nuances and specificities of the brain signals in our study, thereby improving its generalization capability and reducing overfitting \cite{yamashitaConvolutionalNeuralNetworks2018}.

Our architecture is further optimized through differing architecture, such as max-pooling layers. This simplification aids in computational efficiency and reduces the risk of overfitting. Techniques like batch normalization and dropout are also employed to ensure the model's generalizability to new, unseen data.

In summary, our innovative integration of CNNs and CNN-LSTMs into the processing and analysis of ECoG datasets not only achieves higher accuracy rates but also facilitates an end-to-end learning pipeline. This is particularly promising for accelerating the clinical applications of brain-computer interfaces, thereby expanding their potential utility in medical science and beyond.

\section{Results}

In this rigorous investigation, we examined the role of UMAP-based dimensionality reduction and deep learning algorithms, specifically CNNs and CNN-LSTM hybrids, in the analysis of electrocorticography signals for brain-computer interfaces (ECoG BCI). UMAP's effectiveness in classifying ECoG signals appeared to be a good indicator of how well corresponding deep learning models would perform. In other words, when UMAP yielded high classification accuracies, deep learning models also demonstrated excellent performance, particularly in individualized datasets.

\subsection{Dimensionality Reduction Results}

For our UMAP testing we prepared two sets of data: 
\begin{itemize}
    \item Unprocessed and averaged split by participant
    \item Processed and averaged  split by participant
\end{itemize}

For each set, data was randomly split into a training and test set (75:25). In our comprehensive study, we executed UMAP dimensionality reduction on different test-sets to analyze the effectiveness of automated mapping developed on the training set. Observations indicated that datasets localized to individual participants displayed distinct differences between signals, thereby achieving enhanced classification accuracy through the UMAP model\cite{scikit-learn}.

For the individual trials, Individual 3 and 6 was chosen to represent consistent testing parameters and brain location. Then, the data was normalized between 0 and 1 and split into training and test set data, primed for the UMAP modelling step. 

\begin{table}[H]
\centering
\begin{tabular}{lcccc}
\toprule
& Preprocessed Train & Preprocessed Test & No Preprocessed Train & No Preprocessed Test \\
\midrule
0 & 0.8027 & 0.7027 & 0.7959 & 0.7838 \\
1 & 0.9796 & 0.9730 & 0.9388 & 0.9189 \\
2 & 0.7891 & 0.5405 & 0.7483 & 0.7297 \\
3 & 0.9320 & 0.8108 & 0.9796 & 1.0000 \\
4 & 0.9660 & 1.0000 & 0.8571 & 0.8378 \\
5 & 0.9592 & 0.9730 & 1.0000 & 1.0000 \\
6 & 1.0000 & 0.9730 & 1.0000 & 1.0000 \\
\midrule
Avg & 0.9184 & 0.8533 & 0.9028 & 0.8958 \\
\bottomrule
\end{tabular}
\caption{Reformatted KNN Evaluation Results with Train Data Before Test Data}
\label{tab:reformatted-train-before-test}
\end{table}

\subsubsubsection{Unprocessed and Averaged Split by Participant}

\paragraph{Participant Index 3}
\begin{figure}[H]
  \centering
  \includegraphics[width=1\linewidth]{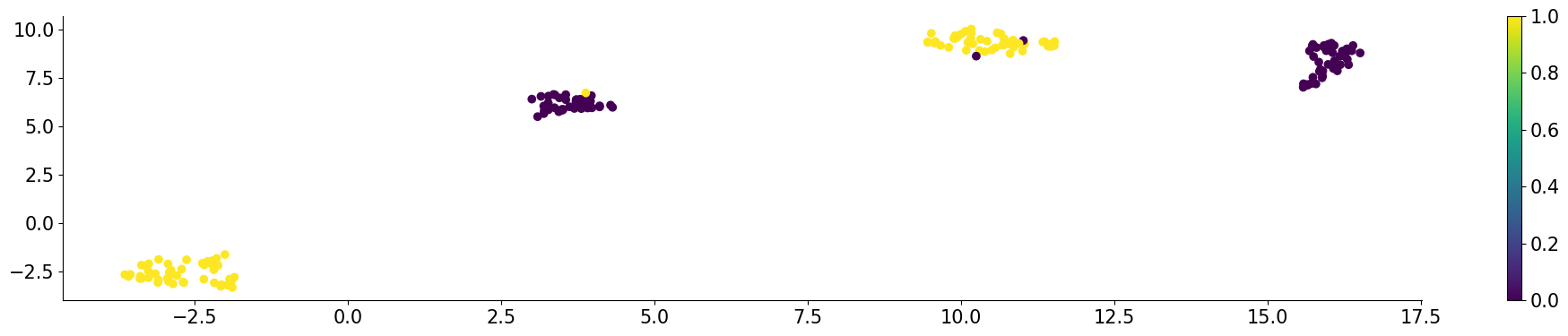}
  \newline
  \includegraphics[width=1\linewidth]{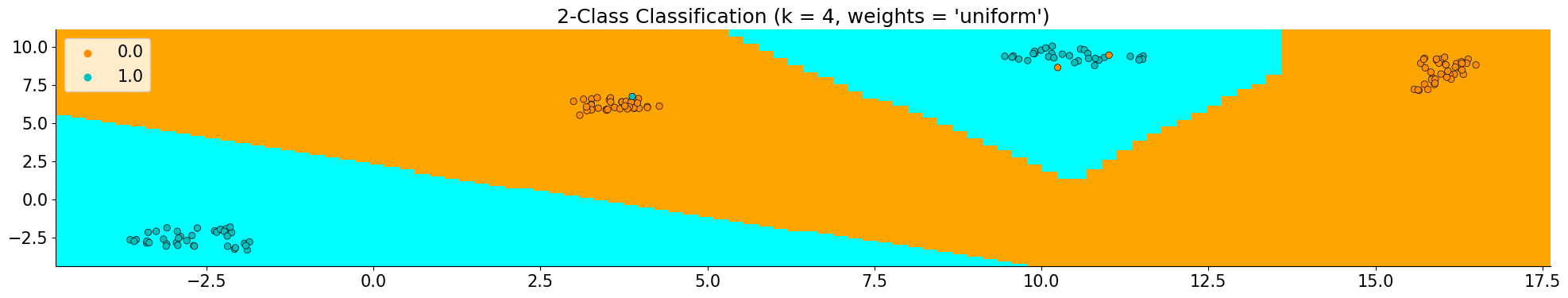}
  \caption{Unprocessed and Averaged UMAP and KNN For Participant 3}
\end{figure}

\noindent Train score: 0.9796, Test score: 1.0000

\paragraph{Participant Index 6}
\begin{figure}[H]
  \centering
  \includegraphics[width=1\linewidth]{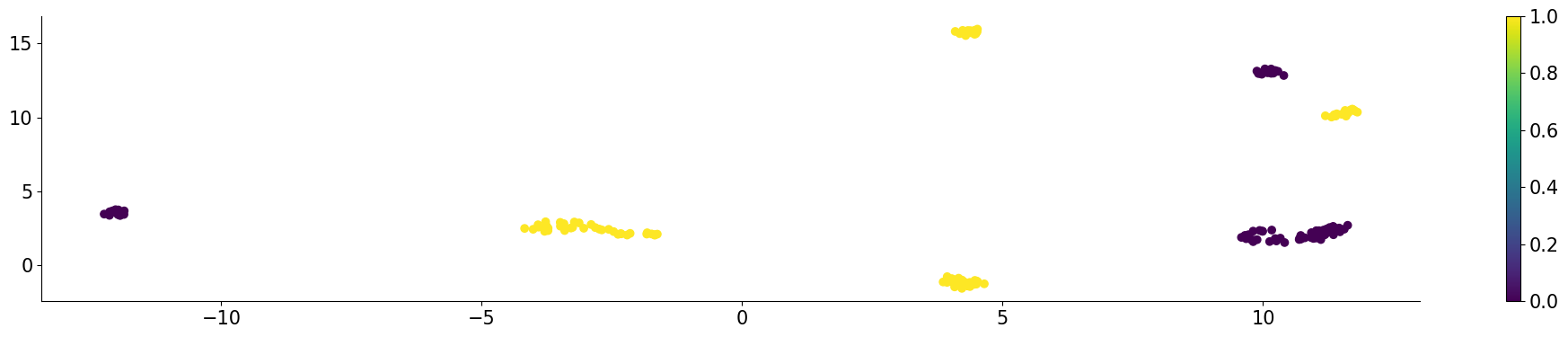}
  \includegraphics[width=1\linewidth]{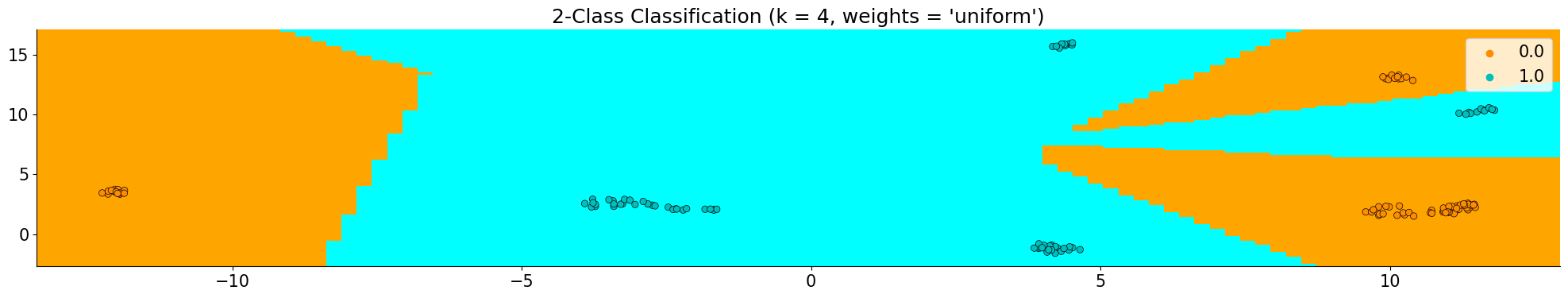}
  \caption{Unprocessed and Averaged UMAP and KNN for Participant Index 6}
\end{figure}

\noindent Train score: 1.0000, Test score: 1.0000

\subsubsubsection{Processed and Averaged Split by Participant}

\paragraph{Participant Index 3}
\begin{figure}[H]
  \centering
  \includegraphics[width=1\linewidth]{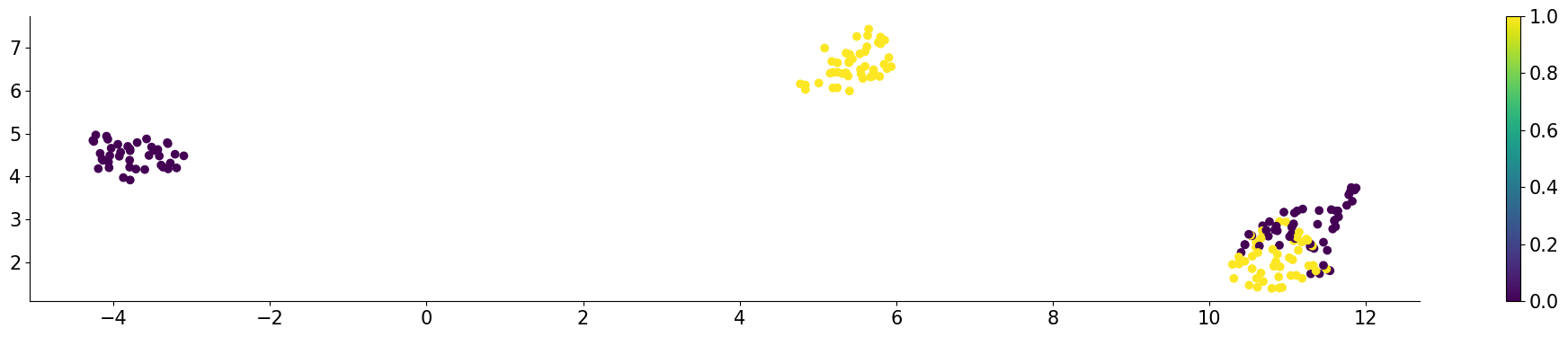}
  \includegraphics[width=1\linewidth]{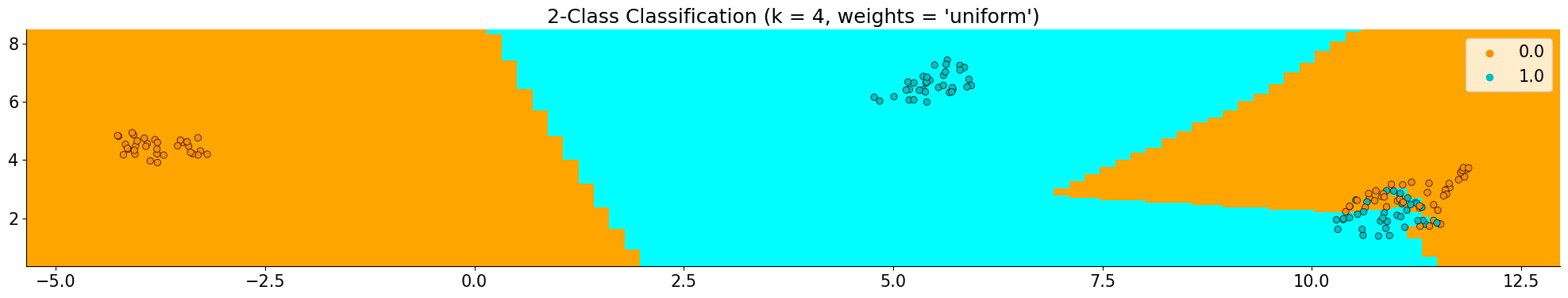}
  \caption{Processed and Averaged UMAP and KNN for Participant Index 3}
\end{figure}

\noindent Train score: 0.9320, Test score: 0.8108

\paragraph{Participant Index 6}
\begin{figure}[H]
  \centering
  \includegraphics[width=1\linewidth]{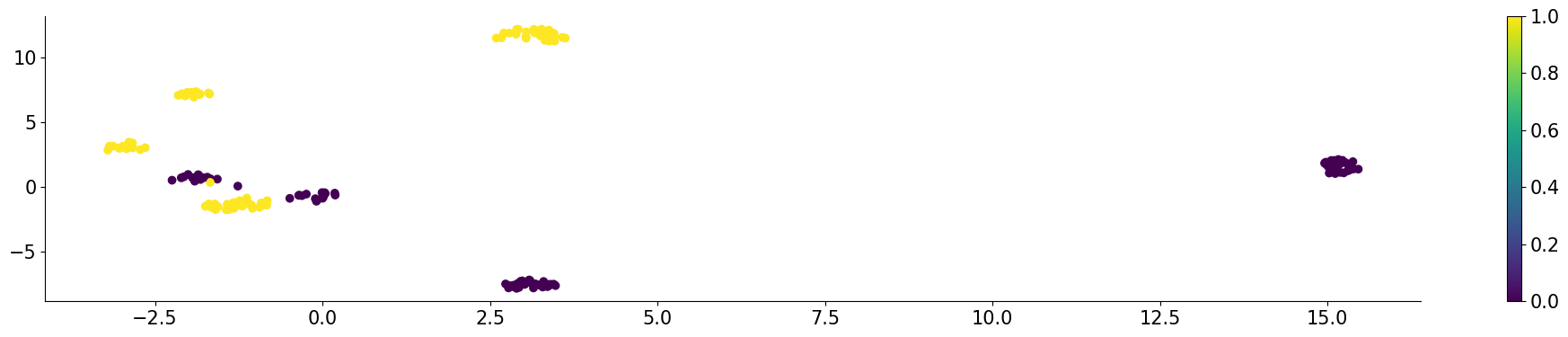}
  \includegraphics[width=1\linewidth]{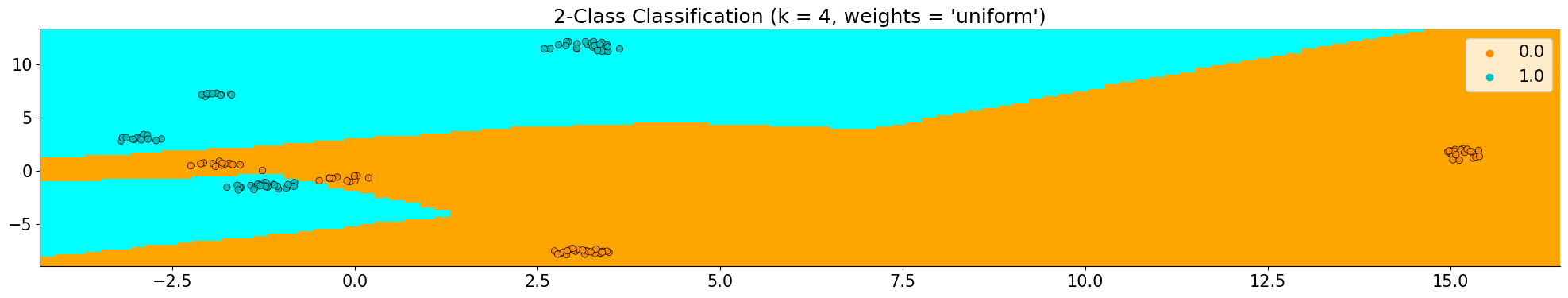}
  \caption{Processed and Averaged UMAP and KNN for Participant Index 6}
\end{figure}

\noindent Train score: 1.0, Test score: 0.9730

These data further substantiate the claim that individual-specific datasets lead to superior signal differentiation and classification through UMAP modeling, reaffirming the relevance of our approach in the BCI context. Images corresponding to each scenario were provided, illustrating the detailed relationships and outcomes for each participant's dataset. Dimensionality reduction proves to be a sound technique in classifying ECoG brain signals.

In this study, we investigated the efficacy of dimensionality reduction techniques, specifically UMAP, in the context of electrocorticography-based brain-computer interfaces (ECoG BCI). Our results demonstrate that UMAP achieves an accuracy of over 85\% without preprocessing and 89.5\% with preprocessing in K-Nearest Neighbors (KNN) tests employing 4 neighbors and uniform weights. Remarkably, the exclusion of data from participant 2 elevates the average KNN accuracy to above 90\%, signifying the influence of individual variances on model performance. 

\subsection{Initial DL Testing}

In our case, we designed two separate CNN and CNN-LSTM model architectures. To first prove that the classification system worked, we tested our models on classifying stimulus states, which were hand stim, tongue stim or resting state\cite{tensorflow2015-whitepaper}.

Initially, we sought to identify key intervals during which specific activities—tongue or hand movements—were recorded. These intervals were determined based on the 't-on' and 't-off' variables within the dataset. To add a layer of complexity, we also isolated the "resting" intervals, which occur between these recorded activities. The intervals for both activity and rest were then combined to create a comprehensive set of time frames, which were subsequently labeled. A '0' or '1' was assigned for tongue and hand activities respectively, while a '2' was designated for resting intervals. Upon segmenting the intervals, we embarked on the cleaning phase, employing a two-step filtering approach to improve the signal-to-noise ratio and capture essential frequency characteristics. This procedure serves to amplify the power associated with different frequency bands within the data, a crucial step for subsequent classification tasks. The data was then normalized by its mean to make the values more comparable across channels.


After these preprocessing steps, the ECoG data for each interval—comprising 3000 samples—was extracted and stored, along with its corresponding label, in NumPy arrays for later use. To set the stage for model training and evaluation, we divided the data into training and test sets in a random 75-25\% ratio. To ensure compatibility with the 2D CNN layers used in the modeling phase, both the training and test sets were reshaped by adding an extra dimension. Additionally, the labels were converted to a one-hot encoded format, making them compatible for categorical cross-entropy loss calculations during the training process. With these steps, our data was fully prepped and primed for effective deep learning-based classification. To further optimize the performance of our neural networks, we employed Keras Tuner, a hyperparameter optimization framework. This allowed us to systematically search for the best hyperparameters for both the CNN and CNN-LSTM models, including aspects such as learning rate, number of filters, and kernel size among others. The tuning process involved running multiple epochs on different configurations and evaluating them based on the test accuracy. This was an essential step in our workflow, ensuring that the models were not only fit for the data but also optimized for generalization, thereby improving their robustness and predictive power. Utilizing Keras Tuner enabled us to fine-tune our models more efficiently, ultimately leading to better-performing models tailored to the specific characteristics of the ECoG data. We searched for the best performing models whilst varying the hyperparameters. 

Ultimately, the best CNN architecture obtained 83\% in test accuracy whilst the best CNN-LSTM, which we hoped could capture more complex features, obtained 73\% accuracy. With this initial modelling complete, we moved on to the real vs imagery based modelling. 

\subsection{DL Real vs. Imagery Modelling}

We applied the same processing steps for the real and imagery motor voltage instances, with a few exceptions. First, the model was only classifying two cases, so the final layer of the model and the labels were changed to reflect that. Second, we restricted the spatial features to the first 48, as not all of the recordings had all 64 spatial locations. After this processing, the data was ready for modelling and hyperparameter tuning. In addition, we selected two participants to attempt to train initial models and test their accuracy. Participant 2 was selected as it had the worst performing accuracies in the KNN and Participant 1 was selected at random from the best performing accuracies. After conducting 30 iterations to fine-tune the model, our best-performing configuration achieved an accuracy of 78\% using a CNN-LSTM architecture and 68\% using a standalone CNN when trained on data from Participant 2. These results were obtained through meticulous hyperparameter optimization, ensuring that the models were both fit for the specific dataset and optimized for generalization.

Initially, fine tuning the model from participant 2 (p2) to participant 1 (p1) to predict p1 data yielded a high test accuracy of 95\%, although further testing is necessary to validate these results. Secondary attempts reached a high test loss of 0.0948 and the test accuracy reached an impressive 98.33\%. This suggests that models benefiting from dimensionality reduction techniques in p2 can seamlessly transition into deep learning tasks in p1, highlighting the discrete nature of the data. However, when a new model was trained on p1 data without fine tuning, the model achieved 100\% accuracy almost immediately. Conversely, applying fine tuning with p1's model on p2's data to predict p2 did not yield comparable results. The model exhibited a test loss of 1.1554 and a significantly lower test accuracy of 46.67\%. This suggests that the model transferability and fine tuning is not symmetric between these two projects. Rather, Dimensionality reduction appears to be a useful metric for determining the complexity of individual brain setups. Generally, if dimensionality reduction techniques are effective, then deep learning models like CNNs and LSTMs can classify the data with high accuracy. This was shown when training and fine tuning hyperparameters on Participant 1, which had high KNN accuracy scores. In a mere 5 trials, the model had already hit 100\% test accuracy. However, if dimensionality reduction is less effective, the chance of accurate classification by CNN and LSTM models drops to around 70\%.

An interesting discrepancy was noted when training participant 4's data on a model from participant 2. While the LSTM-based model only achieved a 58\% accuracy, the CNN-based model performed significantly better with a test accuracy of 78.33\% and a test loss of 0.505. This could be attributed to the LSTM model's potential for overfitting, especially considering that it did not fully account for the time-series nature of the data. When this CNN model was used to train on participant 2's data, it overfitted, evidenced by a test accuracy of 50\% despite a training accuracy of 100\%.To validate these findings, additional tests were conducted on other targets with effective dimensionality reduction, namely Participants 5 and 6. Both participants yielded CNN and LSTM models with test accuracies at or near 100\% after hyperparameter tuning, reinforcing the initial observations regarding the impact of successful dimensionality reduction.

\begin{table}
    \hspace*{-1.5cm} 
    \begin{tabular}{lccccccc}
        \toprule
        Participant & Mean\_Real & Mean\_Imag & Abs\_Mean\_Diff & Std\_Real & Std\_Imag & Range\_Real & Range\_Imag\\
        \midrule
        Alldat      & 0.0268 & 0.0121 & 0.0147 & 0.0096 & 0.0104 & 0.0636 & 0.0547 \\
        Participant 0 & -0.0069 & 0.0140 & 0.0209 & 0.0028 & 0.0103 & 0.0157 & 0.0683 \\
        Participant 1 & 0.0988 & 0.0426 & 0.0562 & 0.0080 & 0.0074 & 0.0539 & 0.0448 \\
        Participant 2 & 0.0144 & 0.0178 & 0.0034 & 0.0097 & 0.0097 & 0.0600 & 0.0709 \\
        Participant 3 & 0.1199 & -0.0498 & 0.1697 & 0.0086 & 0.0087 & 0.0442 & 0.0458 \\
        Participant 4 & 0.0062 & -0.0014 & 0.0076 & 0.0036 & 0.0036 & 0.0207 & 0.0256 \\
        Participant 5 & 0.0779 & 0.0177 & 0.0602 & 0.0108 & 0.0084 & 0.0736 & 0.0530 \\
        Participant 6 & -0.1123 & 0.0244 & 0.1367 & 0.0147 & 0.0146 & 0.0882 & 0.0946 \\
        \bottomrule
    \end{tabular}
    \caption{Fourier and Coherence Analysis Results}
    \label{tab:FourierCoherence}
\end{table}

Quantitative analysis focusing on bootstrapped insights demonstrate interesting results. Particularly, Participants 2 and 4 exhibited the lowest differences between the real and imaginary components of their signals, making them difficult to differentiate. In contrast, other participants demonstrated more substantial differences, potentially accounting for their successful dimensionality reduction and classification. This is quite interesting as every participant besides 2 and 4 that demonstrates differences in coherence of the mean also has relatively high dimensionality reduction accuracy and has high modelling test accuracies. 



\begin{figure}[H]
\hspace*{-1.5cm} 

\includegraphics[width=16cm]{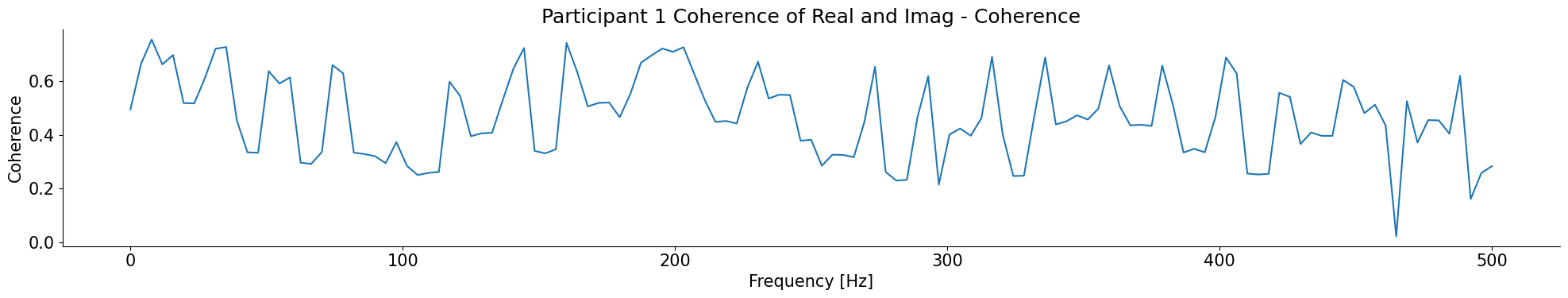}
\caption{Coherence of Participant 1 Imag and Real}
\label{fig:CoherenceP1}
\end{figure}

\begin{figure}[H]
\hspace*{-3cm} 

\includegraphics[width=16cm]{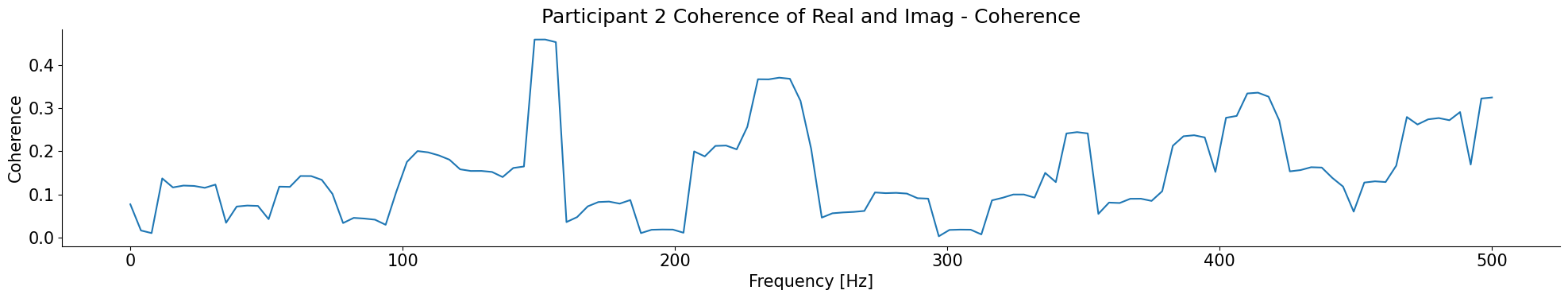}
\caption{Coherence of Participant 2 Imag and Real}
\label{fig:CoherenceP2}
\end{figure}

This is furthered by coherence analysis of Participant 2, who lacked specific frequency spikes that were prevalent in other participants. This absence could explain the difficulties in achieving effective dimensionality reduction and subsequent classification for this participant. Interestingly, Participant 2 exhibited higher coherence between the real and imaginary components of their neural signals compared to Participant 1. While this might suggest a simpler or less noisy neural signal, it's worth noting that Participant 6 also demonstrated high coherence but was successfully classified post-dimensionality reduction.

In alignment with the strong KNN results, our supervised models, including CNNs and LSTMs, generally achieved test accuracies of near 100\%. This high performance is particularly significant given that dimensionality reduction methods, notably UMAP, served as a indicator for DL performance. Interestingly, participants who underperformed in KNN tests also showed lower—but still robust—accuracies in CNN and LSTM models. For instance, Participant 2 registered accuracies of 75\% and 68\% in CNN-LSTM and CNN models, respectively. Further analysis revealed that nearly all participants showing differences in signal coherence also excelled in both dimensionality reduction and supervised learning tasks, underscoring the pivotal role of dimensionality reduction in enhancing model performance within the ECoG BCI context. These composite findings illuminate the intricate relationship between dimensionality reduction techniques, the choice of supervised learning models, and the individual complexities inherent in neural datasets. The study not only validates the effectiveness of our approach but also offers invaluable insights for future advancements in this fast-evolving field.

\section{Discussion}

In this investigation, we delved into the prospects of utilizing deep learning methodologies alongside dimensionality analysis for scrutinizing electrocorticographic (ECoG) data pertinent to brain-computer interfaces (BCIs) centered around motor imagery and execution. Our insights underscore the criticality of tailoring prediction models to individual circumstances and accentuate the role of dimensionality reduction as a foundational step prior to deploying supervised classification techniques on neural data. By employing dimensionality reduction, the necessity for extensive data labeling can be significantly curtailed, thereby conserving time and effort in the supervised facets. This streamlined approach facilitates a more efficient engagement with convolutional neural networks (CNN) and long short-term memory networks (LSTM), resorting to these advanced techniques only in instances where dimensionality reduction falls short in achieving the desired classification accuracy.

\subsection{Project Limitations}
Our analysis faced several limitations due to the dataset used. The dataset featured only a minimal amount of ECoG data and needed to be cut off at 46 brain regions, which may not be equal between recordings. As a result, it is difficult to determine if the models and predictions made with this dataset can be applied to future examples. In addition, neurophysiological responses can vary significantly between individuals due to factors like age, health conditions, or even cognitive states. This variability poses challenges in developing universally effective models.

The takeaways from our study may not transfer over into a broader BCI context because ECoG data tends to be unique. However, the same methodology should hopefully be useful in developing individualized prediction models for other applications. In the future, some methodology is necessary to generalize the classification models because of high inter-subject variabilities in ECoG and EEG signals \cite{huSubjectSeparationNetwork2023}. Currently, high calibration phases in BCI developments make the introduction of new patients long and time consuming, decreasing efficiency. 

The use of deep learning only exacerbates this due to the potential risk of overfitting, which means that models might perform well on the training data but fail to generalize new unseen data. Especially in regards to neural data, deep learning models trained to a specific particiant's dataset cannot generalize to a new participant. In addition, deep learning can suffer from errors in transfer learning (using large datasets to overcome data scarcity), as the gap between conventional datasets and neural signals is vast \cite{huangReviewSignalProcessing2021}.  Thus, the best step to take could be augmenting the existing deep learning paradigm with hand crafted figures and domain adaptation methods that reduce the necessity to capture calibration data of new subjects. \cite{huSubjectSeparationNetwork2023}

Although ECoG datasets are superior in terms of spatial and temporal resolution, it still requires an invasive form of surgery when implanting electrodes onto the brain surface \cite{leuthardtBrainComputerInterface2004}. Any kind of signalling inputs are susceptible to various types of noise and artifacts which can challenge the efficacy of the created deep learning models.  

\subsection{Future Steps}

To further advance our understanding of motor functions in the brain and improve the effectiveness of BCIs, we propose the following future steps:

\begin{itemize}
    \item Augment the existing DL suite with a calibration system that allows for the model to be generalized 
    \item Use the ability of prediction and classification of ECoG data to develop tools that can link imagery and motor execution to create solutions for motor dysfunction
\end{itemize}

\subsection{Human Context and Ethics}

As we continue to explore the potential of BCI informatics in the context of AI, it is essential to address the ethical implications of this technology. BCI technology is in heavy debate, and it is crucial to focus on the regulation of technology as research and development continue. BCI-based technology is already in clinical trials, so it is essential to approach neurotechnology with a specific regulatory framework to ensure research does not infringe on human rights. Research should primarily focus on using BCI technology as a solution for brain dysfunction, such as epilepsy or Parkinson's disease. The rapid pace of BCI research also brings the need for robust regulatory measures to the forefront. In conclusion, our study demonstrates the potential of deep learning techniques in analyzing ECoG data for BCIs focused on motor imagery and execution \cite{drewMindreadingMachinesAre2023}. However, the rapid pace of development necessitates an increased focus on the regulatory, safety, and reliability aspects of this technology. As BCIs become more sophisticated, the need for stringent regulations becomes paramount to ensure the safety and effectiveness of these devices in clinical and everyday use.

\section{Acknowledgements}

Thank you to my mentor, Emma Besier, for helping me through the research and modelling process. 

\section{Disclosure Of Interest}

There is no competing financial interests or personal relationships that would have influenced the work reported in this paper. 

\section{Dataset Availability Statement}

The dataset used in this paper can be found here: \url{https://osf.io/ksqv8/download}, from the Miller 2010 paper\cite{millerCorrectionCorticalActivity2010}. All code used in this research can be found at \url{https://github.com/bafanaS/dim-reduction-with-cnn-lstm.git}

\section{ORCID}

Soham Bafana: \url{ https://orcid.org/0009-0005-3681-1088}
\bibliographystyle{unsrt}    
\bibliography{refs}

\begin{thebibliography}{10}

\bibitem{miaoBCIBasedRehabilitationStroke2020}
Yangyang Miao, Shugeng Chen, Xinru Zhang, Jing Jin, Ren Xu, Ian Daly, Jie Jia, Xingyu Wang, Andrzej Cichocki, and Tzyy-Ping Jung.
\newblock {BCI}-{Based} {Rehabilitation} on the {Stroke} in {Sequela} {Stage}.
\newblock {\em Neural Plasticity}, 2020:1--10, December 2020.

\bibitem{macintyreMotorImageryPerformance2018}
Tadhg~E. MacIntyre, Christopher~R. Madan, Aidan~P. Moran, Christian Collet, and Aymeric Guillot.
\newblock Motor imagery, performance and motor rehabilitation.
\newblock In {\em Progress in {Brain} {Research}}, volume 240, pages 141--159. Elsevier, 2018.

\bibitem{thompsonComputationalLimitsDeep2020}
Neil~C. Thompson, Kristjan Greenewald, Keeheon Lee, and Gabriel~F. Manso.
\newblock The {Computational} {Limits} of {Deep} {Learning}.
\newblock 2020.
\newblock Publisher: arXiv Version Number: 2.

\bibitem{dae-cheolgwonReviewPublicMotor2023}
{Dae-Cheol Gwon}, {K. Won}, {Minseok Song}, {C. Nam}, {S. Jun}, and {M. Ahn}.
\newblock Review of public motor imagery and execution datasets in brain-computer interfaces.
\newblock {\em Frontiers in Human Neuroscience}, 2023.
\newblock S2ID: 76e47bf54fbdb8b4faa120dd9d0b06b669ee23b1.

\bibitem{liDeepLearningEEG2020}
Gen Li, Chang~Ha Lee, Jason~J. Jung, Young~Chul Youn, and David Camacho.
\newblock Deep learning for {EEG} data analytics: {A} survey.
\newblock {\em Concurrency and Computation: Practice and Experience}, 32(18), September 2020.

\bibitem{torresEEGBasedBCIEmotion2020}
Edgar~P. Torres, Edgar~A. Torres, Myriam Hernández-Álvarez, and Sang~Guun Yoo.
\newblock {EEG}-{Based} {BCI} {Emotion} {Recognition}: {A} {Survey}.
\newblock {\em Sensors}, 20(18):5083, September 2020.

\bibitem{Hill2012-pe}
N~Jeremy Hill, Disha Gupta, Peter Brunner, Aysegul Gunduz, Matthew~A Adamo, Anthony Ritaccio, and Gerwin Schalk.
\newblock Recording human electrocorticographic ({ECoG}) signals for neuroscientific research and real-time functional cortical mapping.
\newblock {\em J. Vis. Exp.}, (64), June 2012.

\bibitem{leuthardtBrainComputerInterface2004}
Eric~C Leuthardt, Gerwin Schalk, Jonathan~R Wolpaw, Jeffrey~G Ojemann, and Daniel~W Moran.
\newblock A brain–computer interface using electrocorticographic signals in humans.
\newblock {\em Journal of Neural Engineering}, 1(2):63--71, June 2004.

\bibitem{volkovaDecodingMovementElectrocorticographic2019}
Ksenia Volkova, Mikhail~A. Lebedev, Alexander Kaplan, and Alexei Ossadtchi.
\newblock Decoding {Movement} {From} {Electrocorticographic} {Activity}: {A} {Review}.
\newblock {\em Frontiers in Neuroinformatics}, 13:74, December 2019.

\bibitem{raviprakashDeepLearningProvides2020}
Harish RaviPrakash, Milena Korostenskaja, Eduardo~M. Castillo, Ki~Hyeong Lee, Ki~Hyeong Lee, Christine~M. Salinas, James~E. Baumgartner, James Baumgartner, {Syed Muhammad Anwar}, Syed~Muhammad Anwar, {Syed Muhammad Anwar}, Concetto Spampinato, and Ulas Bagci.
\newblock Deep {Learning} {Provides} {Exceptional} {Accuracy} to {ECoG}-{Based} {Functional} {Language} {Mapping} for {Epilepsy} {Surgery}.
\newblock {\em Frontiers in Neuroscience}, 14:409--409, May 2020.
\newblock MAG ID: 3016238703 S2ID: 18645282f0e1d036cddca2f12f114f4c525a6584.

\bibitem{wangDecodingSemanticInformation2011}
Wei Wang, Alan~D. Degenhart, Gustavo Sudre, {Gustavo Sudre}, Dean~A. Pomerleau, and Elizabeth~C. Tyler-Kabara.
\newblock Decoding semantic information from human electrocorticographic ({ECoG}) signals.
\newblock {\em Annual International Conference of the IEEE Engineering in Medicine and Biology Society}, 2011:6294--6298, January 2011.
\newblock MAG ID: 2011652684 S2ID: 22f42bf9cad4a49e6adf30e8e9036f8b7dbe098b.

\bibitem{krizhevskyImageNetClassificationDeep2017}
Alex Krizhevsky, Ilya Sutskever, and Geoffrey~E. Hinton.
\newblock {ImageNet} classification with deep convolutional neural networks.
\newblock {\em Communications of the ACM}, 60(6):84--90, May 2017.

\bibitem{devlinBERTPretrainingDeep2018}
Jacob Devlin, Ming-Wei Chang, Kenton Lee, and Kristina Toutanova.
\newblock {BERT}: {Pre}-training of {Deep} {Bidirectional} {Transformers} for {Language} {Understanding}.
\newblock 2018.
\newblock Publisher: arXiv Version Number: 2.

\bibitem{mcinnesUMAPUniformManifold2018}
Leland McInnes, John Healy, and James Melville.
\newblock {UMAP}: {Uniform} {Manifold} {Approximation} and {Projection} for {Dimension} {Reduction}.
\newblock 2018.
\newblock Publisher: arXiv Version Number: 3.

\bibitem{kimConvolutionalNeuralNetwork2017}
Phil Kim.
\newblock Convolutional {Neural} {Network}.
\newblock In {\em {MATLAB} {Deep} {Learning}}, pages 121--147. Apress, Berkeley, CA, 2017.

\bibitem{duDecodingECoGSignal2018}
Anming Du, Yang Shuqin, Liu Weijia, and Haiping Huang.
\newblock Decoding {ECoG} {Signal} with {Deep} {Learning} {Model} {Based} on {LSTM}.
\newblock {\em IEEE Region 10 Conference}, pages 430--435, October 2018.
\newblock MAG ID: 2918941196 S2ID: 0a399e9a9ce9c1d97bd44e4d7aa64f43a165e8b1.

\bibitem{garza-ulloaDeepLearningModels2022}
Jorge Garza-Ulloa.
\newblock Deep {Learning} {Models} {Evolution} {Applied} to {Biomedical} {Engineering}.
\newblock In {\em Applied {Biomedical} {Engineering} {Using} {Artificial} {Intelligence} and {Cognitive} {Models}}, pages 509--607. Elsevier, 2022.

\bibitem{sliwowskiDeepLearningECoG2022}
Maciej Śliwowski, Matthieu Martin, Antoine Souloumiac, Pierre Blanchart, and Tetiana Aksenova.
\newblock Deep learning for {ECoG} brain-computer interface: end-to-end vs. hand-crafted features.
\newblock 2022.
\newblock Publisher: arXiv Version Number: 2.

\bibitem{millerCorrectionCorticalActivity2010}
Kai~J. Miller, Gerwin Schalk, Eberhard~E. Fetz, Marcel~den Nijs, Jeffrey~G. Ojemann, and Rajesh P.~N. Rao.
\newblock Correction for {Cortical} activity during motor execution, motor imagery, and imagery-based online feedback,.
\newblock {\em Proceedings of the National Academy of Sciences of the United States of America}, 107(15), January 2010.
\newblock MAG ID: 2522353246 S2ID: f43332961277074544e3e43ae6dae5d9763d9e30.

\bibitem{harris2020array}
Charles~R. Harris, K.~Jarrod Millman, St{\'{e}}fan~J. van~der Walt, Ralf Gommers, Pauli Virtanen, David Cournapeau, Eric Wieser, Julian Taylor, Sebastian Berg, Nathaniel~J. Smith, Robert Kern, Matti Picus, Stephan Hoyer, Marten~H. van Kerkwijk, Matthew Brett, Allan Haldane, Jaime~Fern{\'{a}}ndez del R{\'{i}}o, Mark Wiebe, Pearu Peterson, Pierre G{\'{e}}rard-Marchant, Kevin Sheppard, Tyler Reddy, Warren Weckesser, Hameer Abbasi, Christoph Gohlke, and Travis~E. Oliphant.
\newblock Array programming with {NumPy}.
\newblock {\em Nature}, 585(7825):357--362, September 2020.

\bibitem{scikit-learn}
F.~Pedregosa, G.~Varoquaux, A.~Gramfort, V.~Michel, B.~Thirion, O.~Grisel, M.~Blondel, P.~Prettenhofer, R.~Weiss, V.~Dubourg, J.~Vanderplas, A.~Passos, D.~Cournapeau, M.~Brucher, M.~Perrot, and E.~Duchesnay.
\newblock Scikit-learn: Machine learning in {P}ython.
\newblock {\em Journal of Machine Learning Research}, 12:2825--2830, 2011.

\bibitem{huSubjectSeparationNetwork2023}
Haochen Hu, Kang Yue, Mei Guo, Kai Lu, and Yue Liu.
\newblock Subject {Separation} {Network} for {Reducing} {Calibration} {Time} of {MI}-{Based} {BCI}.
\newblock {\em Brain Sciences}, 13(2):221, January 2023.

\bibitem{vicentea.lomelin-ibarraMotorImageryAnalysis2022}
{Vicente A. Lomelin-Ibarra}, {Andres E. Gutierrez-Rodriguez}, and Jose~Antonio Cantoral-Ceballos.
\newblock Motor {Imagery} {Analysis} from {Extensive} {EEG} {Data} {Representations} {Using} {Convolutional} {Neural} {Networks}.
\newblock {\em Sensors}, 22(16):6093--6093, August 2022.
\newblock MAG ID: 4291824710 S2ID: 90c316e44e74cc4e62068aa95b40e3db6fdf159b.

\bibitem{wangUnsupervisedDecodingLongTerm2016}
Nancy Xin~Ru Wang, Jared~D. Olson, Jeffrey~G. Ojemann, Rajesh P.~N. Rao, and Bingni~W. Brunton.
\newblock Unsupervised {Decoding} of {Long}-{Term}, {Naturalistic} {Human} {Neural} {Recordings} with {Automated} {Video} and {Audio} {Annotations}.
\newblock {\em Frontiers in Human Neuroscience}, 10:165--165, April 2016.
\newblock ARXIV\_ID: 1511.08260 MAG ID: 2236184431 S2ID: 7cf8440b1c02c021f6ba8543ad490b4788bbe280.

\bibitem{liMotorImageryEEG2022}
Hongli Li, Man Ding, Ronghua Zhang, and Chunbo Xiu.
\newblock Motor imagery {{EEG}} classification algorithm based on {{CNN-LSTM}} feature fusion network.
\newblock {\em Biomedical Signal Processing and Control}, 72:103342, February 2022.

\bibitem{yamashitaConvolutionalNeuralNetworks2018}
Rikiya Yamashita, Mizuho Nishio, Richard Kinh~Gian Do, and Kaori Togashi.
\newblock Convolutional neural networks: an overview and application in radiology.
\newblock {\em Insights into Imaging}, 9(4):611--629, August 2018.

\bibitem{tensorflow2015-whitepaper}
Mart\'{i}n Abadi, Ashish Agarwal, Paul Barham, Eugene Brevdo, Zhifeng Chen, Craig Citro, Greg~S. Corrado, Andy Davis, Jeffrey Dean, Matthieu Devin, Sanjay Ghemawat, Ian Goodfellow, Andrew Harp, Geoffrey Irving, Michael Isard, Yangqing Jia, Rafal Jozefowicz, Lukasz Kaiser, Manjunath Kudlur, Josh Levenberg, Dandelion Man\'{e}, Rajat Monga, Sherry Moore, Derek Murray, Chris Olah, Mike Schuster, Jonathon Shlens, Benoit Steiner, Ilya Sutskever, Kunal Talwar, Paul Tucker, Vincent Vanhoucke, Vijay Vasudevan, Fernanda Vi\'{e}gas, Oriol Vinyals, Pete Warden, Martin Wattenberg, Martin Wicke, Yuan Yu, and Xiaoqiang Zheng.
\newblock {TensorFlow}: Large-scale machine learning on heterogeneous systems, 2015.
\newblock Software available from tensorflow.org.

\bibitem{huangReviewSignalProcessing2021}
Xin Huang, Yilu Xu, Jing Hua, Wenlong Yi, Hua Yin, Ronghua Hu, and Shiyi Wang.
\newblock A {Review} on {Signal} {Processing} {Approaches} to {Reduce} {Calibration} {Time} in {EEG}-{Based} {Brain}-{Computer} {Interface}.
\newblock {\em Frontiers in Neuroscience}, 15:733546, 2021.

\bibitem{drewMindreadingMachinesAre2023}
Liam Drew.
\newblock Mind-reading machines are coming — how can we keep them in check?
\newblock {\em Nature}, 620(7972):18--19, August 2023.

\end{thebibliography}
\end{document}